\pdfoutput=1
\documentclass[10pt, conference, compsocconf]{IEEEtran}

\usepackage{pbalance}

\usepackage{upquote}

\usepackage[ngerman,main=english]{babel}
\addto\extrasenglish{\languageshorthands{ngerman}\useshorthands{"}}

\usepackage[hyphens]{url}

\makeatletter
\g@addto@macro{\UrlBreaks}{\UrlOrds}
\makeatother

\usepackage[zerostyle=b,scaled=.75]{newtxtt}

\usepackage[T1]{fontenc}

\usepackage[
  babel=true, % Enable language-specific kerning. Take language-settings from the languge of the current document (see Section 6 of microtype.pdf)
  expansion=alltext,
  protrusion=alltext-nott, % Ensure that at listings, there is no change at the margin of the listing
  final % Always enable microtype, even if in draft mode. This helps finding bad boxes quickly.
]{microtype}

\DisableLigatures{encoding = T1, family = tt* }

\usepackage{graphicx}
\usepackage{subfigure}

\usepackage{tikz}
\usepackage{color}
\usepackage{floatrow}    
\usetikzlibrary{arrows, calc, positioning, shapes.geometric, backgrounds, fit}

\usepackage{diagbox}

\usepackage{xcolor}

\usepackage{listings}

\definecolor{eclipseStrings}{RGB}{42,0.0,255}
\definecolor{eclipseKeywords}{RGB}{127,0,85}
\colorlet{numb}{magenta!60!black}

\lstdefinelanguage{json}{
    basicstyle=\normalfont\ttfamily,
    commentstyle=\color{eclipseStrings}, % style of comment
    stringstyle=\color{eclipseKeywords}, % style of strings
    numbers=left,
    numberstyle=\scriptsize,
    stepnumber=1,
    numbersep=8pt,
    showstringspaces=false,
    breaklines=true,
    frame=lines,
    string=[s]{"}{"},
    comment=[l]{:\ "},
    morecomment=[l]{:"},
    literate=
        *{0}{{{\color{numb}0}}}{1}
         {1}{{{\color{numb}1}}}{1}
         {2}{{{\color{numb}2}}}{1}
         {3}{{{\color{numb}3}}}{1}
         {4}{{{\color{numb}4}}}{1}
         {5}{{{\color{numb}5}}}{1}
         {6}{{{\color{numb}6}}}{1}
         {7}{{{\color{numb}7}}}{1}
         {8}{{{\color{numb}8}}}{1}
         {9}{{{\color{numb}9}}}{1}
}

\usepackage[autostyle=true]{csquotes}

\defineshorthand{"`}{\openautoquote}
\defineshorthand{"'}{\closeautoquote}

\usepackage{booktabs}

\usepackage{paralist}

\usepackage[%
  square,        % for square brackets
  comma,         % use commas as separators
  numbers,       % for numerical citations;
  sort&compress % as sort but in addition multiple numerical citations
]{natbib}

\usepackage{etoolbox}
\makeatletter
\patchcmd{\NAT@test}{\else \NAT@nm}{\else \NAT@hyper@{\NAT@nm}}{}{}
\makeatother

\usepackage{pdfcomment}

\usepackage{stfloats}
\fnbelowfloat

\usepackage[group-minimum-digits=4,per-mode=fraction]{siunitx}
\addto\extrasgerman{\sisetup{locale = DE}}

\usepackage[incolumn]{mindflow}

\usepackage[capitalise,nameinlink,noabbrev]{cleveref}

\crefname{listing}{Listing}{Listings}
\Crefname{listing}{Listing}{Listings}
\crefname{lstlisting}{Listing}{Listings}
\Crefname{lstlisting}{Listing}{Listings}

\usepackage{lipsum}

\usepackage[math]{blindtext}
\usepackage{mwe}
\usepackage[realmainfile]{currfile}
\usepackage{tcolorbox}
\tcbuselibrary{listings}

\DeclareFontFamily{U}{MnSymbolC}{}
\DeclareSymbolFont{MnSyC}{U}{MnSymbolC}{m}{n}
\DeclareFontShape{U}{MnSymbolC}{m}{n}{
  <-6>    MnSymbolC5
  <6-7>   MnSymbolC6
  <7-8>   MnSymbolC7
  <8-9>   MnSymbolC8
  <9-10>  MnSymbolC9
  <10-12> MnSymbolC10
  <12->   MnSymbolC12%
}{}
\DeclareMathSymbol{\powerset}{\mathord}{MnSyC}{180}

\usepackage{xspace}

\expandafter\def\csname ver@subfig.sty\endcsname{}
\makeatletter
\newcommand{\hydash}{\penalty\@M-\hskip\z@skip}
\makeatother

\input glyphtounicode
\begin{document}
% Enable following command if you need to typeset "IEEEpubid".
% See https://bytefreaks.net/tag/ieeeoverridecommandlockouts for details.
%\IEEEoverridecommandlockouts

\title{CANDID: Correspondence AligNment for Deep-burst Image Denoising}

\author{%
  \IEEEauthorblockN{Arijit Mallick\IEEEauthorrefmark{1}\textsuperscript{\textsection}}
  \IEEEauthorblockA{Department of Computer Graphics\\University of T\"ubingen, Germany \\
    \url{arijit.mallick@uni-tuebingen.de}}    
  \and
   \IEEEauthorblockN{Raphael Braun}
  \IEEEauthorblockA{Department of Computer Graphics\\University of T\"ubingen, Germany \\
    \ \url{raphael.braun@uni-tuebingen.de}}
    \and
   \IEEEauthorblockN{Hendrik PA Lensch}
  \IEEEauthorblockA{Department of Computer Graphics\\University of T\"ubingen, Germany \\
    \ \url{hendrik.lensch@uni-tuebingen.de}}

}

%%%%%%%%%%%%%% TEASER %%%%%%%%%%%%%%%%%%%%%%%

\begin{comment}

\author{%
  \IEEEauthorblockN{Anonymous submission}
  \IEEEauthorblockA{Paper ID 7}
}
\makeatletter
\let\@oldmaketitle\@maketitle% Store \@maketitle
\renewcommand{\@maketitle}{\@oldmaketitle% Update \@maketitle to insert...
\centering
\includegraphics[width=\linewidth,height=6\baselineskip]
    {example-image}
     \caption{teaser}
     \label{fig:CANDID}
     \bigskip}% ... an image
\makeatother
\end{comment}

\maketitle
% In case you want to add a copyright statement.
% Works only in the compsoc conference mode.
%
% Source: https://tex.stackexchange.com/a/325013/9075
%
% All possible solutions:
%  - https://tex.stackexchange.com/a/325013/9075
%  - https://tex.stackexchange.com/a/279134/9075
%  - https://tex.stackexchange.com/q/279789/9075 (TikZ)
%  - https://tex.stackexchange.com/a/200330/9075 - for non-compsocc papers

\iffalse
  \IEEEoverridecommandlockouts
  \IEEEpubid{\begin{minipage}{\textwidth}\ \\[12pt] \centering
      1551-3203 \copyright 2015 IEEE.
      Personal use is permitted, but republication/redistribution requires IEEE permission.
      \\
      See \url{https://www.ieee.org/publications_standards/publications/rights/index.html} for more information.
    \end{minipage}}
\fi

\begingroup\renewcommand\thefootnote{\textsection}
\footnotetext{This work was done while Mr. Arijit Mallick was affiliated with University of T\"ubingen, Germany. He is currently working for TWT GmbH Science \& Innovation, Germany.}
\endgroup

%\title{My awesome paper}

\begin{abstract}
With the advent of mobile phone photography and point-and-shoot cameras, deep-burst imaging is widely used for a number of photographic effects such as depth of field, super-resolution, motion deblurring, and image denoising. 
In this work, we propose to solve the problem of deep-burst image denoising by including an optical flow-based correspondence estimation module which aligns all the input burst images with respect to a reference frame. In order to deal with varying noise levels the individual burst images are pre-filtered with different settings. Exploiting the established correspondences one network block predicts a pixel-wise spatially-varying filter kernel to smooth each image in the original and prefiltered bursts before fusing all images to generate the final denoised output. The resulting pipeline achieves state-of-the-art results by combining all available information provided by the burst. 
\end{abstract}

% \fi
\IEEEpeerreviewmaketitle

%-------------------------------------------------------------------------
\section{Introduction}
\label{sec:intro}
Due to recent development of faster and lightweight portable CPUs especially in mobile and point-and-shoot cameras, burst photography has been gaining further prominence because of the noise reduction and motion blur removal capabilities. Burst photography~\cite{sr2,sr3} can also be understood as multi-frame image restoration task \cite{Liba_2019,burst_p1,bhat1,burst_sr1} which has a wider range of applications even in satellite photography~\cite{valsesia2021piunet,highresnet2020} 
for remote sensing.
The sensors and lenses in smartphones are much smaller and more lightweight than those of professional cameras but they collect less light per pixel which leads to noisier images. Compensating this by longer exposures could introduce motion blur. 
As an alternative, a burst of many short and noisy images could be computationally combined into one sharp image. Camera phone APIs already provide denoising algorithms but they are not optimized for burst denoising.

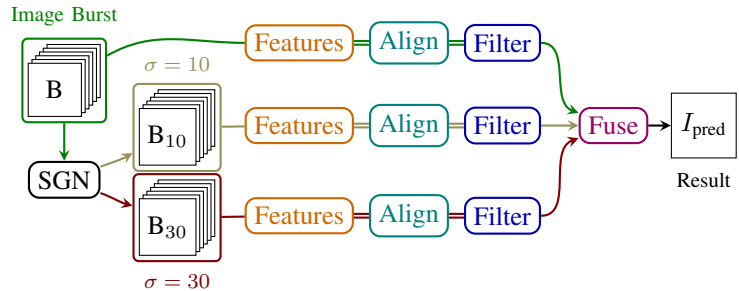
\begin{figure*}[htb]
\centering
    \begin{tikzpicture}[square/.style={regular polygon,regular polygon sides=4}]
    % input image stack
    \foreach \i in {0,...,5} {
        \node[draw, fill, fill=white, square, inner sep = 0pt, minimum width=1cm] (input_\i) at(-\i*0.05,-\i*0.05 + 0.125) {B};
    }
    \scoped[on background layer]{\node [fit=(input_0)(input_5), draw=green!50!black, inner sep=2pt, rounded corners=2pt, thick] (input_box) {};}
    \node[color=green!50!black, above = 1pt of input_box] {\footnotesize Image Burst};
    % SGN
    \node[draw, rounded corners, thick, below=5mm of input_box] (sgn) {SGN};
    
    % denoised image stacks
    \foreach \i in {0,...,5} {
        \node[draw, fill, fill=white, square, inner sep = 0pt, minimum width=1cm] (sgn_10_\i) at(1.5-\i*0.05,-\i*0.05 -0.5) {B$_{10}$};
    }
    \scoped[on background layer]{\node [fit=(sgn_10_0)(sgn_10_5), draw=yellow!50!black, inner sep=2pt, rounded corners=2pt, thick] (sgn_10_box) {};}
    \node[color=yellow!50!black, above = 1pt of sgn_10_box] {\footnotesize $\sigma = 10$};
    \foreach \i in {0,...,5} {
        \node[draw, fill, fill=white, square, inner sep = 0pt, minimum width=1cm] (sgn_30_\i) at(1.5-\i*0.05,-\i*0.05 -1.7) {B$_{30}$};
    }
    \scoped[on background layer]{\node [fit=(sgn_30_0)(sgn_30_5), draw=red!50!black, inner sep=2pt, rounded corners=2pt, thick] (sgn_30_box) {};}
    \node[color=red!50!black, below = 1pt of sgn_30_box] {\footnotesize $\sigma = 30$};
    
    % Feature Extraction
    \node[draw, thick, rounded corners, color=orange!80!black] (features_input) at (3.0, 0.5) {Features};
    \node[draw, thick, rounded corners, color=orange!80!black] (features_sgn_10) at (3.0, -0.6) {Features};
    \node[draw, thick, rounded corners, color=orange!80!black] (features_sgn_30) at (3.0, -1.8) {Features};
    
    % Alignment
    \node[draw, thick, rounded corners, color=blue!50!green, right = 2mm of features_input] (align_input) {Align};
    \node[draw, thick, rounded corners, color=blue!50!green, right = 2mm of features_sgn_10] (align_sgn_10) {Align};
    \node[draw, thick, rounded corners, color=blue!50!green, right = 2mm of features_sgn_30] (align_sgn_30)  {Align};
    
    % Spatial filtering
    \node[draw, thick, rounded corners, color=blue!70!black, right = 2mm of align_input] (filter_input) {Filter};
    \node[draw, thick, rounded corners, color=blue!70!black, right = 2mm of align_sgn_10] (filter_sgn_10) {Filter};
    \node[draw, thick, rounded corners, color=blue!70!black, right = 2mm of align_sgn_30] (filter_sgn_30) {Filter};
    
    % combination
    \node[draw, thick, rounded corners, color=purple!80!blue, right = 5mm of filter_sgn_10] (combine){Fuse};
    
    % Result
    \node[draw, fill, fill=white, square, right =3mm of combine, inner sep = 0pt, minimum width=1cm] (result) {$I_{\text{pred}}$};   
    \node[below = 2pt of result] {\footnotesize Result};

    % Arrows and paths
    \path[draw, thick, -stealth, green!50!black] (input_box) to [out=-90, in=90] (sgn);
    
    \path[draw, yellow!50!black, thick, -stealth] (sgn) to[out=20,in=210] (sgn_10_box);
    \path[draw, red!50!black, thick, -stealth] (sgn) to[out=-20,in=165] (sgn_30_box);
    
    \path[draw, thick, green!50!black] (input_box) to[out=30, in=180] (features_input);
    \path[draw, thick, double, green!50!black] (features_input) to (align_input) to (filter_input);
    \path[draw, thick, -stealth, green!50!black] (filter_input) to[out=0, in=160] (combine);
    
    \path[draw, thick, yellow!50!black] (sgn_10_box) to (features_sgn_10);
    \path[draw, thick, double, yellow!50!black] (features_sgn_10) to (align_sgn_10) to (filter_sgn_10);
    \path[draw, thick, -stealth, yellow!50!black](filter_sgn_10) to [out=0, in=180] (combine);
    
    \path[draw, thick, red!50!black] (sgn_30_box) to (features_sgn_30);
    \path[draw, thick, double, red!50!black] (features_sgn_30) to (align_sgn_30) to (filter_sgn_30);
    \path[draw, thick, -stealth, red!50!black] (filter_sgn_30) to [out=0, in=200] (combine);
    
    \path[draw, thick, -stealth] (combine) to (result);

\end{tikzpicture}
    \vspace*{-10mm}
    \caption{Method overview.
    The input \textcolor{green!50!black}{image burst} is pre-filtered twice using SGN \cite{gu2019self} with different filter strengths.
    For each stack we extract \textcolor{orange!80!black}{features}, \textcolor{blue!50!green}{align} both features and images and then apply a content-adaptive spatial \textcolor{blue!80!black}{filter} with weights derived from the aligned features.
    The results from all three bursts are \textcolor{purple!80!blue}{fused} to predict the denoised output.}%
    \label{fig:method_overview}
\end{figure*}

Current burst denoising methods perform the alignment simply based on homographies estimated between the burst images~\cite{Godard_2018_ECCV,bhat2021deep}, followed by some pixel denoising.
Our proposed pixel-wise alignment based on optical flow is significantly more powerful in compensating for scenes with complex depth and camera or object motion. We still expect the motion between the burst images to be rather small. 
%Unlike expensive cost-volume-based registration methods, our iterative pixel correspondence estimation network consists only of a U-net and MLPs. Based on U-net-based features that encode the local neighborhood the MLPs predict the offset to the potential corresponding pixel that depicts the same scene point. Starting with the same pixel location in the reference frame, i.e.\ the center frame of the burst, and the secondary image this offset is iteratively refined by always considering the U-net features of the current estimate. The process converges quickly. 
Our overall architecture is depicted in Figure~\ref{fig:method_overview}.
First, we generate enhanced burst inputs by applying the pre-trained self-guided filtering network (SGN)~\cite{gu2019self} for each image to generate pre-denoised bursts. 
Both, the original and the denoised images are aligned with respect to the reference frame using the RAFT ~\cite{teed2020raft} optical flow network. Based on the aligned images and extracted features a network block predicts a per-pixel adaptive filter kernel to denoise every pixel in every image. A final fusion block merges all predictions across all bursts into a single output.

Secondly, we estimate pixel adaptive filter-kernels which per pixel describe where to collect color information from the aligned input bursts. The decoder then only applies those kernels, thus produces weighted averages over neighbouring pixels from all aligned images.

We demonstrate the importance of each module in an ablation study. Our contributions are as follows: a) optical flow-based alignment of multiple pre-denoised burst images, b) adaptive per-pixel filtering of aligned burst images followed by cross-burst fusion, c) improved denoising performance, especially in low-noise scenarios.

%-------------------------------------------------------------------------

\section{Related Work}
\label{sec:related}
Related work covers single image denoising, homography-based alignment, and deep-burst imaging.
In the following section, we discuss related works pertaining to our problem statement, starting with single image denoising, followed by homography-based and optical flow-based alignment, and finally contemporary progress on deep-burst imaging.\\
%------------------------------------------------------------------------- 
\paragraph{Single image denoising}
%Image denoising is a classical computer vision problem and is still one of the most sought after deep learning based low-level image processing research topics. 
%Due to the increasing popularity of low-cost mobile photography, effective denoising and enhancement is well sought after. 
Most photography hardware companies take advantage of the recently developed lightweight neural network denoising models; exploiting the significant increase in mobile computation power. In the early days of CNNs, models such as~\cite{gu2019self} improved performance compared to classical image denoising models based on Markov random fields but they could not compete with BM3D~\cite{BM3D} which introduced a new denoising paradigm by combining 3D block matching and domain transform.  
They are later surpassed by a sparse denoising autoencoder models~\cite{inpainting,KSVD}. 
Simple multi-layer perceptron-based models~\cite{fields}, residual link networks~\cite{resD} and later deeper residual networks~\cite{imRestore} and persistent memory-based networks~\cite{Memnet} have shown superior performance due to enhanced receptive fields. All these models have the advantage of being trainable end-to-end exploiting simple to generate training data.
For a multitude of image processing tasks, training can be accelerated using pre-trained models and transfer learning~\cite{Itrans}.
In this spirit, we incorporate the pre-trained self-guided network (SGN)~\cite{gu2019self} to enrich the burst input with smooth priors. SGN extracts large-scale contextual information and gradually propagates it to the higher resolution sub-networks for feature self-guidance and denoising at multiple scales. This efficient multiscale local features extraction property allows it to efficiently recover denoised images.

\paragraph{Deep-burst Denoising}
While single image denoising relies on learned image priors, deep-burst denoising assimilates features from multiple noisy frames to predict a better image. A similar idea is used in burst motion deblurring~\cite{Wieschollek_2017_ICCV} where a sharp image is recurrently extracted from a burst of blurry ones. Similarly, recurrent neural networks have also been used for burst denoising. Bhat et al.\ \cite{bhat2021deep} reparametrize the image formation process in the latent space, and integrate learned image priors for the denoised prediction.
Kernel prediction networks~\cite{mildenhall2018kpn,mkpn} leverage the localized pixel neighborhood weighted filters to predict a denoised image from multiple inputs.  Dudhane et al.\cite{bipnet22} proposed to extract pre-processed features from each burst frame following an edge-boosting burst alignment module. The pseudo-burst features are then enriched using multi-scale contextual information, which is followed by adaptively aggregate information from the corresponding features.

Our novel burst denoising model also applies adaptive pixel-neighborhood filters but first performs an explicit alignment step.

%Our novel burst denoising model is an upgraded, alignment-based extension of multi-frame alignment plus a decoder denoising architecture. 
%------------------------------------------------------------------------- 

\paragraph{Correspondence Alignment}
 Multiple frame denoising usually involves some sort of alignment~\cite{bhat1}
 of the frames in the burst for superior feature assimilation. Tico~\cite{mfid} demonstrates a block matching approach within the reference and the neighboring frames to support multiple frame denoising. VBM4D~\cite{VBM4D} and VBM3D~\cite{VBM3D} take the BM3D algorithm further to video denoising with faster homography flow-based alignment. We instead estimate per pixel correspondences for a more fine-grained alignment. 
When capturing a burst of images of a potentially dynamic scene with a handheld camera each image will show slightly different content. 
In order to effectively utilize information from those multiple frames for denoising, the frames need to be aligned~\cite{bhat1}.

%For that purpose, we propose a novel alignment module with the help of pixel-wise iterative dense correspondence matching. 
Tico~\cite{mfid} demonstrates a block matching approach within the reference and the neighboring frames to support multiple frame denoising.
VBM4D~\cite{VBM4D} and VBM3D~\cite{VBM3D} take the BM3D algorithm further to video denoising with block matching for alignment. 

%Handcrafted keypoints detectors are generally robust to domain changes but are more time-consuming to craft than their learnable counterparts. Additionally,
%Strong scene changes, however, severely affect the performance of the handcrafted methods, while
Neural network optical flow models can leverage information beyond patch-level correspondence information to predict dense correspondences, i.e.\ estimating pixel motion between consecutive frames of a video \cite{bruhn2005lucas}.
Some of the first learning based optical flow methods used simple CNN architectures \cite{buades2016, dosovitskiy2015flownet, ilg2017flownet}. Recently they were superseded by recurrent techniques like RAFT~\cite{teed2020raft} or transformer-based architectures like FlowFormer~\cite{huang2022flowformer}. Those current state of the art techniques are very good and very close to ground truth \cite{Butler:ECCV:2012}.

%Contrary to the aforementioned methods, we encode the pixel neighborhood in both, the reference and every secondary image in such a way that a multi-layer perception (MLP) can predict the distance vector to the true corresponding pixel given the encoded features of the reference pixel and an arbitrary pixel in its vicinity. As the quality of the predictions depends on how close the secondary pixel is to the true correspondence we iterate this process. 
 In our approach we utilize the success in the optical flow field by using a pretrained RAFT implementation provided in torchvision~\cite{NEURIPS2019_9015}.
 RAFT provides the high-quality pixel-wise correspondence alignment that we rely on for our denoising approach.

%-------------------------------------------------------------------------

\section{Method}
The core idea of our burst denoising method is to first spatially align the pixels in the burst stack. 
Afterwards, each aligned image is denoised by a content-adaptive spatially-varying filter step followed by an adaptive fusion of all processed images (see Figure~\ref{fig:method_overview}, left). 
%the color information of pixels over the burst is combined for denoising in a content adaptive filtering step over the aligned images.
%An overview of the method architecture is shown in Figure~\ref{fig:method_overview}.

\subsection{Prefiltering with SGN}\label{sec:prefilter}
\begin{figure}[htb]
    \centering
     \subfigure[Input]{\includegraphics[height=.4\textwidth]{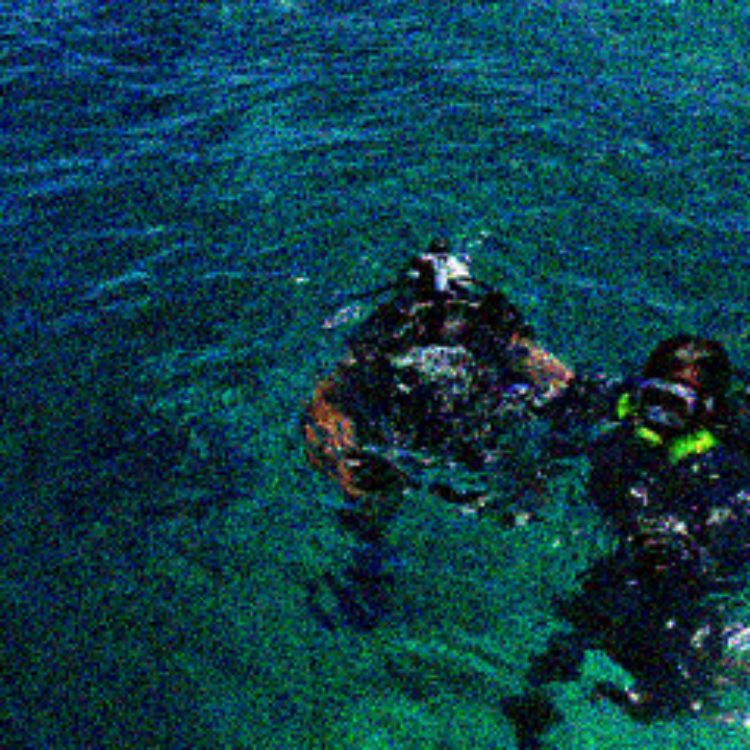}}
    %\hspace{.8cm}
    \subfigure[Ground Truth]{\includegraphics[height=.4\textwidth]{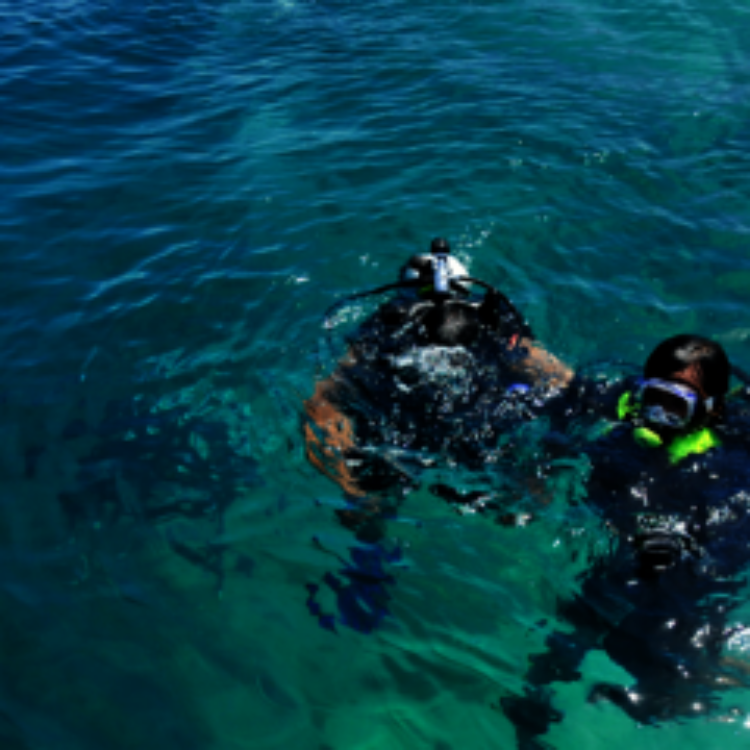}}
    \subfigure[SGN $\sigma=10$] {\includegraphics[height=.4\textwidth]{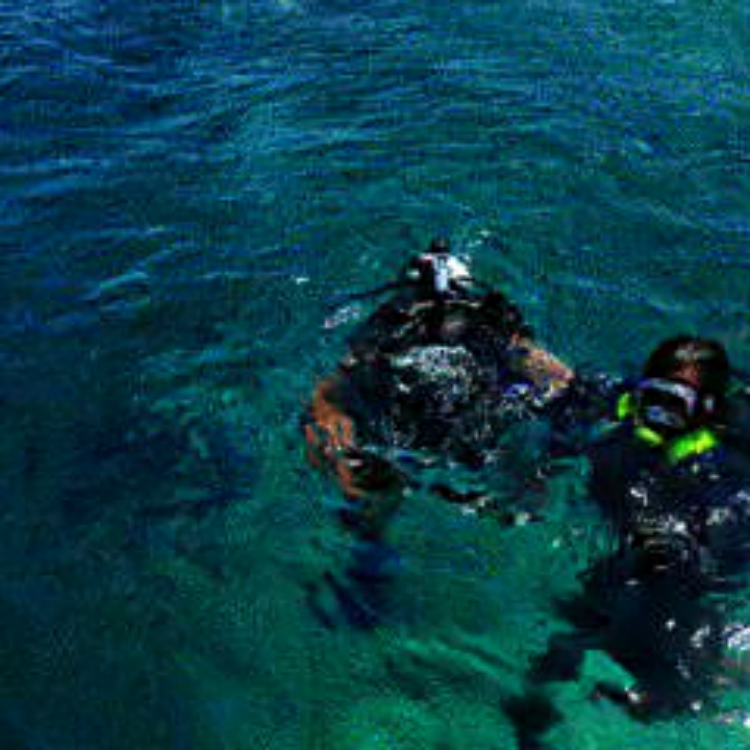}} 
    %\hspace{.8cm}
    \subfigure[SGN $\sigma=30$]{\includegraphics[height=.4\textwidth]{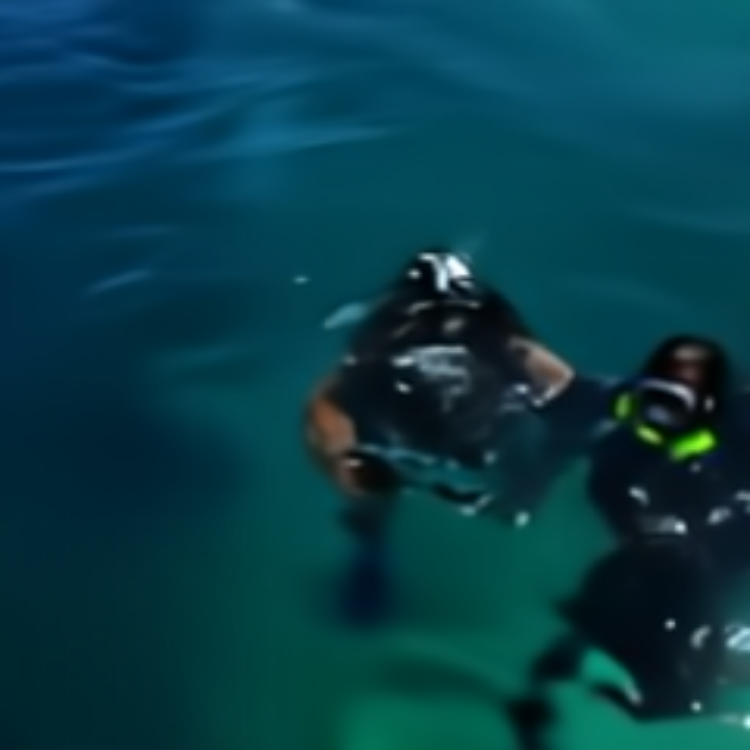}} \\[-.5ex]
    {\small Denoising performance of the pre-trained SGN. ($\sigma=10$) retains the original sharpness. ($\sigma=30$) shows better denoising performance but loses sharp details.} \\[-1ex]
    \caption{Prefiltering with SGN}%
    \label{fig:SGN_performance}
\end{figure}

Our method starts by filtering the burst images.
The amount of noise in input bursts can vary significantly, even within the same burst.
Because of varying degrees of noise and blur due to abrupt camera motion, precise alignment might be difficult. 
We, therefore, duplicate the input burst into three processing streams.
The first stream B uses the original burst, the second stream B$_{10}$ ($\sigma=10$) a mildly pre-denoised version of the burst and the last one B$_{30}$ ($\sigma=30$) a strongly denoised version (see Figure~\ref{fig:SGN_performance}).
%For this task, any single-frame denoising algorithm could be applied to each individual frame.
We apply the pretrained SGN \cite{gu2019self} to each individual frame but any single-frame denoising algorithm could be used.
%The effect of both denoising levels is shown in Figure~\ref{fig:SGN_performance}.
The intermediate results from the different streams will be fused in the last step of our pipeline.
%We demonstrate the effectiveness of this decision in an ablation study.

\begin{figure}[htb]
   
    \centering
    \vspace{0.2cm}
    \subfigure[Reference]{\includegraphics[height=.3\textwidth]{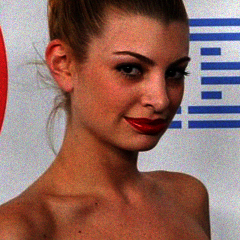}} %\\[-1ex]
    \subfigure[before alignment]{\includegraphics[height=.3\textwidth]{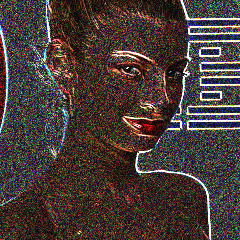}}
    %\hspace{.8cm}
    \subfigure[after alignment]{\includegraphics[height=.3\textwidth]{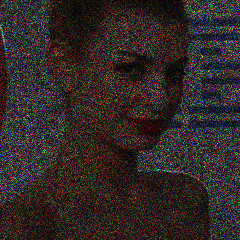}}
    \\[-0.5 ex]
    \raggedright
     {\small Alignment error between the reference and the last burst image scaled 5 times. Note, how after alignment differences in the silhouette are no longer present.} \\[-1ex]
    \vspace*{0.3cm}
    \caption{Pixel-wise alignment}%
    \label{fig:alignment_example}
\end{figure}

\subsection{Feature Extraction}
To add local context to each pixel we enrich each image by processing it with a simple CNN.
In addition, the estimated noise level of the image is concatenated as the fourth channel before processing.
In each processing stream, we produce corresponding feature stacks.
The same shared weights are used for each image in each stream.
Both the image stack and the feature stack are used as inputs for the alignment module.
% \begin{figure}[tb]
%     \centering
%     \include{images/model_feature}
%     \vspace*{-10mm} 
%     \caption{The \textcolor{orange!80!black}{feature} extraction module is a CNN that is applied to each image individually. It produces feature maps that are depicted by the red arrow. The blue arrow shows the unchanged image, which is also used as input to the alignment module.}
%     \label{fig:feature_extractor}
% \end{figure}

\subsection{Alignment}
The central property that is exploited with burst denoising is that the content captured in the individual frames of the burst is very similar.
In the original images, the scene content however might be shifting due to camera shake or scene dynamics.
We use the pre-trained RAFT \cite{teed2020raft} model that is shipped with torchvision \cite{NEURIPS2019_9015} to estimate the optical flow between the reference image frame and any other image frame in the burst. The estimated flow computed from the reference and secondary images is used to warp the secondary image frames and their corresponding feature maps with respect to the reference image frame and the reference feature frame respectively.
The effectiveness of the RAFT-based alignment is visualized in Figure~\ref{fig:alignment_example}.

\subsection{Collaborative Content-adaptive Spatial Filtering}
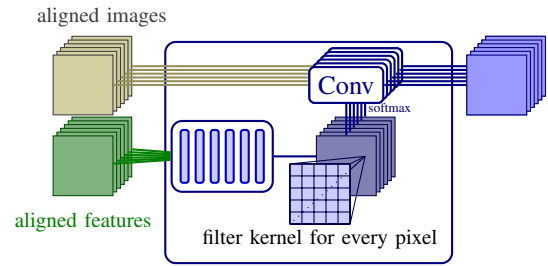
\begin{figure}[htb]
   \centering
   \begin{tikzpicture}[square/.style={regular polygon,regular polygon sides=4}]
%input images
\foreach \i in {0,...,5} {
    \node[draw, yellow!10!black, fill=yellow!60!black!50, square, inner sep=8pt] (input_images_\i) at(-1.5 -\i*0.05,-\i*0.05 + 1.2) {};
}
\node[above=0mm of input_images_0, yellow!10!black] {\footnotesize aligned images};
%input feature stack
\foreach \i in {0,...,5} {
    \node[draw, green!30!black, fill=green!40!black!50, square, inner sep=8pt] (feature_\i) at(-1.5 -\i*0.05,-\i*0.05 + 0.125) {};
}

% cnn
\foreach \i in {0,...,5} {
    \node[minimum width=3pt, minimum height = 20, rounded corners=1pt, draw, thick, color=blue!80!black, inner sep=0pt, fill=blue!20] (cnn_\i)
    at (-0.4+\i/5, 0) {};
}

\scoped[on background layer]{\node [fit=(cnn_0)(cnn_5),
draw=blue!50!black, inner sep=3pt, rounded corners, thick] (cnn) {};}

% output kernel stack
\foreach \i in {0,...,5} {
    \node[draw, blue!30!black, fill=blue!40!black!50, square, inner sep=8pt] (kernel_\i) at(2 -\i*0.05,-\i*0.05 + 0.125) {};
    \node[draw, rounded corners=2pt, color=blue!50!black, thick, fill=white] (filter_\i) at (2 - \i*0.05, -\i*0.05 + 1.2) {Conv};
}
\node[blue!50!black] at (2.33, 0.65) {\tiny softmax};

\foreach \i in {0,...,5} {
    \node[draw, blue!50!black, fill=blue!40, square, inner sep=8pt] (result_\i) at(4.0 -\i*0.05,-\i*0.05 + 1.2) {};
}

% zoom in
\coordinate (pixel) at (2, 0);
\path[draw, blue!30!black] (pixel) to (1, -0.9);
\path[draw, blue!30!black] (pixel) to (1, -0.1);
\path[draw, blue!30!black] (pixel) to (1.8, -0.1);
\path[draw, blue!30!black] (pixel) to (1.8, -0.9);
\path[draw, blue!30!black, step=0.8/5, xshift=1.0cm, yshift=-0.9cm, fill=blue!20!white] (0, 0) grid (0.8, 0.8) rectangle (0, 0);
\path[draw, blue!30!black, dotted] (pixel) to (1, -0.9);
\node (label) at (1.4, -1.1) {\footnotesize filter kernel for every pixel};

\begin{scope}[on background layer]
\path[draw, thick, blue!50!black] (cnn) to ($(cnn.east) + (0.745, 0)$);
\end{scope}
\foreach \i in {0,...,5} {
    \path[draw, thick, green!50!black] (feature_\i) to (cnn);
    \path[draw, thick, yellow!50!black] (input_images_\i) to (filter_\i);
    \path[draw, thick, blue!50!black] (kernel_\i) to (filter_\i);
    \path[draw, thick, blue!50!black] (filter_\i) to[out=0, in=180] (result_\i);
    \node[draw, rounded corners=2pt, color=blue!50!black, thick, fill=white] (filter_\i) at (2 - \i*0.05, -\i*0.05 + 1.2) {Conv};
    \node[draw, blue!50!black, fill=blue!40, square, inner sep=8pt] (result_\i) at(4.0 -\i*0.05,-\i*0.05 + 1.2) {};
}

\scoped[on background layer]{\node [fit=(cnn)(filter_0)(label),
draw=blue!50!black, inner sep=2pt, rounded corners, thick] (cnn) {};}

\node[below=1mm of feature_5, green!50!black] {\footnotesize aligned features};

\end{tikzpicture}
   \vspace*{-1cm}
   \caption{The spatial content-adaptive filter kernels for every pixel are estimated by a CNN based on all \textcolor{green!50!black}{aligned features}.
        They are applied individually to the \textcolor{yellow!50!black}{aligned images} to produce the \textcolor{blue!90!black}{spatially filtered burst}.}
   \label{fig:spatial_filter}
\end{figure}

At this point, the images and features in the bursts are all aligned with respect to the reference frame. The next step is to filter the images spatially and combine the results pixel-wise for the final result. The spatial filtering is implemented with content-dependent per-pixel kernels. Those kernels are estimated by a CNN from the aligned feature stack, i.e. collaboratively considering all feature maps at the same time. The output activations of this CNN are reshaped into $3\times3$ and $5\times5$ filter kernels for all images and all pixels.
The result is two kernels of shape $[N, H, W, 3, 3]$ and $[N, H, W, 5, 5]$ with the number of images $N$, height $H$ and width $W$. The kernels are normalized via \textit{softmax} and applied to each image in the burst individually,  effectively computing a weighted average color over the $3\times3$ and $5\times5$ neighborhood of each pixel as shown in Figure~\ref{fig:spatial_filter}.

\subsection{Burst Fusion}
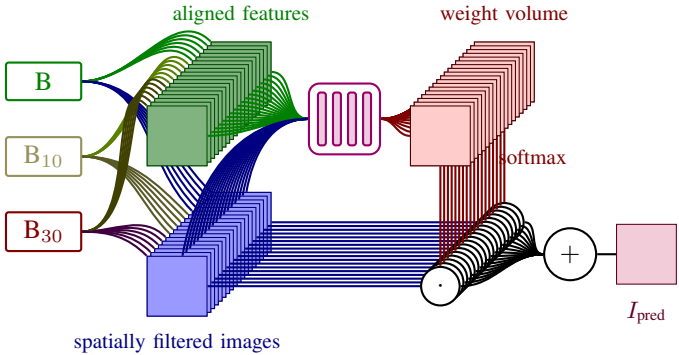
\begin{figure}[htb]
   \centering
   \begin{tikzpicture}[square/.style={regular polygon,regular polygon sides=4}]

\coordinate (in_1) at (-2.5, 1);
\coordinate (in_2) at (-2.5, 0);
\coordinate (in_3) at (-2.5, -1);
\node[thick, draw, rounded corners=1pt, minimum width=1cm, minimum height=5mm, green!50!black] (in_1) at (in_1) {B};
\node[thick, draw, rounded corners=1pt, minimum width=1cm, minimum height=5mm, yellow!50!black] (in_2) at (in_2) {B$_{10}$};
\node[thick, draw, rounded corners=1pt, minimum width=1cm, minimum height=5mm, red!50!black] (in_3) at (in_3) {B$_{30}$};
\foreach \i in {0,...,17} {
    \node[draw, green!30!black, fill=green!40!black!50, square, inner sep=8pt] (input_features_\i) at(0.125 -\i*0.05,0.125-\i*0.05 + 1) {};
    \node[draw, blue!50!black, fill=blue!40, square, inner sep=8pt] (input_images_\i) at(0.125 -\i*0.05,0.125-\i*0.05 -1) {};
}

\foreach \i in {0,...,5} {
    \path[draw, thick, green!50!black] (in_1) to[out=0, in=135] (input_features_\i);
    \path[draw, thick, blue!50!black] (in_1) to[out=0, in=135] (input_images_\i);
}
\foreach \i in {6,...,11} {
    \path[draw, thick, green!30!yellow!50!black] (in_2) to[out=0, in=135] (input_features_\i);
    \path[draw, thick, blue!30!yellow!50!black] (in_2) to[out=0, in=135] (input_images_\i);
}
\foreach \i in {12,...,17} {
    \path[draw, thick, green!50!red!50!black] (in_3) to[out=0, in=135] (input_features_\i);
    \path[draw, thick, blue!50!red!50!black] (in_3) to[out=0, in=135] (input_images_\i);
}
% repaint boxes on top
\foreach \i in {0,...,17} {
    \node[draw, blue!50!black, fill=blue!40, square, inner sep=8pt] (input_images_\i) at(0.125 -\i*0.05,0.125-\i*0.05 -1) {};
    \node[draw, green!30!black, fill=green!40!black!50, square, inner sep=8pt] (input_features_\i) at(0.125 -\i*0.05,0.125-\i*0.05 + 1) {};
}
\node[above =1mm of input_features_0, green!50!black] {\footnotesize aligned features};
\node[below =1mm of input_images_17, blue!50!black] {\footnotesize spatially filtered images};

% cnn
\foreach \i in {1,...,4} {
    \node[minimum width=3pt, minimum height = 20, rounded corners=1pt, draw, thick, color=purple!80!blue, inner sep=0pt, fill=purple!80!blue!20] (cnn_\i)
    at (1.5-0.5+\i/5, 0.5) {};
}
\scoped[on background layer]{\node [fit=(cnn_1)(cnn_4),
draw=purple!80!blue, inner sep=3pt, rounded corners, thick] (cnn) {};}
% cnn result 
\foreach \i in {0,...,17} {
    \path[draw, thick, green!50!black] (input_features_\i) to[out=0, in=180] (cnn);
    \path[draw, thick, blue!50!black] (input_images_\i) to[out=80, in=180] (cnn);
    \node[draw, red!30!black, fill=red!20, square, inner sep=8pt] (weights_\i) at(3.5+0.125 -\i*0.05,0.125-\i*0.05 + 1) {};
    \path[draw, thick, red!50!black] (cnn) to[out=0, in=180] (weights_\i);
}
\node[above =1mm of weights_0, red!50!black] {\footnotesize weight volume};
\node[red!50!black] at(4, 0) {\footnotesize softmax};

% multiply
\foreach \i in {0,...,17} {
    \node[draw, thick, circle, fill=white] (multiply_\i) at (3.5+0.125 -\i*0.05, 0.125-\i*0.05 -1) {$\cdot$};
    \path[draw, thick, blue!50!black] (input_images_\i) to (multiply_\i);
    \path[draw, thick, red!50!black] (weights_\i) to[out=-90, in=90] (multiply_\i);
}
% redraw cnn results
\foreach \i in {0,...,17} {
    \node[draw, red!30!black, fill=red!20, square, inner sep=8pt] (weights_\i) at(3.5+0.125 -\i*0.05,0.125-\i*0.05 + 1) {};
}

% sum
\node[circle, draw, thick] (sum) at (4.5, -1.3) {$+$};
\foreach \i in {0,...,17} {
    \path[draw, thick] (multiply_\i) to[out=0, in=180] (sum);
}

% result
\node[draw, purple!50!black, fill=purple!20, square, inner sep=8pt, right = 0.25cm of sum] (result) {};
\path[draw, thick] (sum) to[out=0, in=180] (result);
\node[below =1mm of result, purple!50!black] {\small $I_{\text{pred}}$};

\end{tikzpicture}
   \vspace*{-1cm}
   \caption{Fusion Network. The \textcolor{green!50!black}{aligned features} from all three bursts are concatenated with the \textcolor{blue!90!black}{spatially filtered images} and processed by a \textcolor{purple!50!black}{CNN} to obtain a \textcolor{red!50!black}{weight volume}.
        The weights are used to compute the \textcolor{purple!50!black}{denoised result} as a weighted per-pixel sum over the \textcolor{blue!90!black}{spatially filtered images}.}
   \label{fig:fusion}
\end{figure}

Remember in Section~\ref{sec:prefilter} the burst was split into three processing streams B, B$_{10}$ and B$_{30}$, which are all processed individually in the same way so far.
This means at this stage we have aligned and spatially filtered images and the corresponding aligned image features for each stream.
The final step is to fuse all information from the different bursts into a single denoised image $I_{\text{pred}}$.
This denoised result is computed as a weighted average over the spatially filtered images from all three processing streams.
As indicated in Figure~\ref{fig:fusion}, we concatenate the aligned features of the streams with the spatially filtered images and process them together in a 4-layer CNN. This CNN produces the weight volume.
This volume contains a weight for every pixel and color of every image.
A softmax over the image dimension is applied to the weights in order to ensure that summing up the weights over this dimension yields 1 for every color channel.
The result $I_{\text{pred}}$ is finally computed as a weighted sum per pixel. This is implemented as element-wise multiplication between weight volume and spatially filtered images followed by a sum over the burst dimension.
Every channel for every input image is therefore weighted individually, which is more powerful than just mixing the existing colors of the spatially filtered images.

\subsection{Training}
Some components of the denoising pipeline  like the SGN and the alignment module are pretrained.
We stopped the gradients from going through the SGN networks, which effectively turns the three burst streams B, B$_{10}$ and B$_{30}$ into separate inputs.
The RAFT network in the alignment module was frozen and used as fixed differentiable operation.
The remaining trainable weights are in the CNNs for the feature extractor, the content-adaptive cooperative spatial filter, and the burst fusion module.
We train end-to-end with ADAM \cite{kingma2014adam} from a simple $L1$-loss $ \mathcal{L} = \left\|I_{\text{pred}} - I_{\text{gt}}\right\|_1$ on the ground truth $I_{gt}$.
% \begin{align*}
%     \mathcal{L} &= \left\|I_{\text{pred}} - I_{\text{gt}}\right\|_1
% \end{align*}

%\hlnote{might add $L1$-loss for individual denoised burst image after content-adaptive filtering -- if it actually helps.}
%\rbnote{Well it does help for low noise levels (lvl.1). The results for higher noise levels are significantly worse however.}
%\begin{figure}[htb]
%\centering
%\subcaptionbox{Input reference}{\includegraphics[width=0.23\textwidth]{images/results/neucor_ref.png}}%
%\hfill % <-- Seperation
%\subcaptionbox{Before alignment(it=0)}{\includegraphics[width=0.23\textwidth]{images/results/neucor_0x5.png}}%
%\hfill % <-- Seperation
%\subcaptionbox{After alignment(it=5)}{\includegraphics[width=0.23\textwidth]{images/results/neucor_5x5.png}}%

%\caption{ For a given input burst reference frame, we compute error maps between the ground truth and the last burst image. Error map pixels are magnified 5 times for a better visual understanding of the alignment abilities. One can clearly notice the alignment artefacts disappear due to our correspondence alignment module and the remaining error is primarily due to additive Gaussian noise.  }
%\end{figure}

%===========================================================
\section{Experiments}
We evaluate our method by comparing to state of the art and validate our architecture choices with an ablation study.
\subsection{Training and experimental setup} 
For the pre-denoising we use the SGN pre-trained with $\sigma=10,30$ \cite{SGNgit}. 
%The neural correspondence alignment has been trained wit for around 24 hours using 4 GPUs.
For the burst denoising training both the SGN pre-denoising and the RAFT alignment model are frozen.
We trained on the OpenImages~\cite{OpenImages2} dataset and evaluated on the grayscale burst benchmark~\cite{mildenhall2018kpn} and RGB burst benchmark following the usual conventions \cite{mkpn,bhat2021deep,bipnet22}.
The ground truth images are shifted and corrupted by adding heteroscedastic Gaussian noise~\cite{hetero} with variance $\sigma^2_r + \sigma^2_s x$.
Here $x$ is the clean pixel value, while $\sigma_r$ and $\sigma_s$ denote the readout and shot noise parameters, respectively. 
Those noise parameters are assumed to be known both during training and testing, and are used in the feature extractor. 
During training they are sampled uniformly in the log-domain from the range $\log(\sigma_r) \in [-3, -1.5]$ and $\log(\sigma_s) \in [-4, -2]$.
The comparisons are evaluated with 2 different noise lvl.1 and lvl.2, corresponding to noise parameters (-2.2, -2.6) and (-1.8, -2.2) respectively.
Training was done on 2 TITAN Xp GPUs and took about 96 hours to converge. %For details on the networks please refer to the Supplemental Materials.
%------- TODO Clean up and move to supplementals ------------------------------------------- 
%\subsection{Hyperparameters}
%
%We use a batch size of 4, learning rate of 1e-3 and the Adam optimizer on the iterative neural correspondence pre-training. The U-net encodes the input images into 27 channel features and has XX LSTM blocks \hlnote{first mentioning of LSTM blocks???} once the corresponding value has been sampled form the given query coordinate. This followed by a separate coordinate XX layer deep MLP with 256 perceptron dimension throughout the architecture. We use 5 iterations during training and testing. 
%%for obtaining the correct the alignment during scenenetRGBD validation as well as burst alignment dataset.
%Additionally, we select 8192 sample points from reference images with 16 randomly generated points around each sample reference points with a standard deviation of 5. (TODO: exact sampling formula). 
%We use a 12 layers deep U-net for burst feature extraction, which shares weights along all the bursts. Hence, given the burst size of $N=8$, we have 24 images and corresponding feature maps in total. Following convention\cite{bhat2021deep}, we take 128x128 image patches randomly sampled from 2000 images. Our feature dimension for the main denoising network is kept at 128 and it takes around 8 hours of training to converge for a batch size of 4 in a multi-GPU setup. We take the shared \hlnote{shared???} filter kernel of kernel size 5 before the prediction network. 

%------------------------------------------------------------------------- 
\subsection{Results}
The quantitative comparison with other methods in Table~\ref{tab:quantitative} shows that our model delivers overall state-of-the-art performance the aforementioned benchmark in the evaluation of the model-unseen dataset.
On deep introspection, we can say that due to SGN and further multiple kernel-based filtering, the model successfully recovers the image even from the heavy noise scenarios.
%This is the same in low noise cases except that we achieve second best results in the very low noise scenario (Lvl.\ 1) as the SGN filtering might have excessively smoothed out the finer details of the original image to compensate for extremely high noise case scenarios.
In the future, one could add additional SGN-based denoising stages with different pre-trained noises to analyze whether further boosting of \textit{lvl.1} and \textit{lvl.2} would be possible. Additionally, larger filter kernels can be added to the model in order to enhance the results for higher noise scenarios. 
%by retraining the network with a standard deviation of 5 or less so that the network gets additional information regarding handling very low noise cases.
%Nevertheless, our model performs second best in terms of PSNR by a small margin in low noise cases while achieving state-of-the-art in intermediate as well as high noise scenarios with overall superiority. 
%------------------------------------------------------------------------- 
%\subsection{Qualitative Results}

\begin{figure*}
   \centering
\begin{tabular}{ccccc}

\includegraphics[height=.2\textwidth, width = .18\textwidth ]{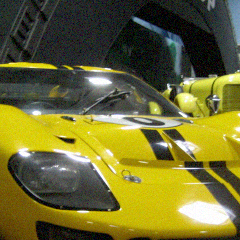}&
\includegraphics[height=.2\textwidth, width = .18\textwidth ]{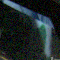}&
\includegraphics[height=.2\textwidth, width = .18\textwidth ]{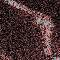}&
\includegraphics[height=.2\textwidth, width = .18\textwidth ]{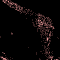}&
\includegraphics[height=.2\textwidth, width = .18\textwidth ]{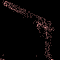}\\

\includegraphics[height=.2\textwidth, width = .18\textwidth ]{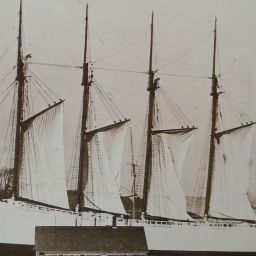}&
\includegraphics[height=.2\textwidth, width = .18\textwidth ]{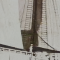}&
\includegraphics[height=.2\textwidth, width = .18\textwidth ]{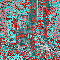}&
\includegraphics[height=.2\textwidth, width = .18\textwidth ]{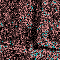}&
\includegraphics[height=.2\textwidth, width = .18\textwidth ]{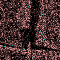}\\

\includegraphics[height=.2\textwidth, width = .18\textwidth ]{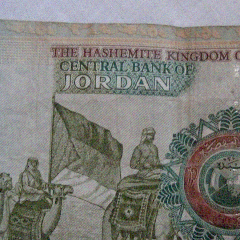}&
\includegraphics[height=.2\textwidth, width = .18\textwidth ]{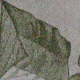}&
\includegraphics[height=.2\textwidth, width = .18\textwidth ]{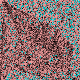}&
\includegraphics[height=.2\textwidth, width = .18\textwidth ]{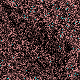}&
\includegraphics[height=.2\textwidth, width = .18\textwidth ]{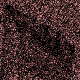}\\

\includegraphics[height=.2\textwidth, width = .18\textwidth ]{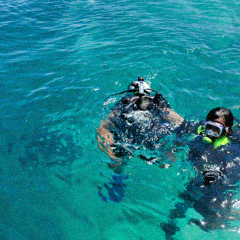}&
\includegraphics[height=.2\textwidth, width = .18\textwidth ]{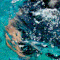}&
\includegraphics[height=.2\textwidth, width = .18\textwidth ]{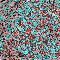}&
\includegraphics[height=.2\textwidth, width = .18\textwidth ]{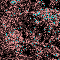}&
\includegraphics[height=.2\textwidth, width = .18\textwidth ]{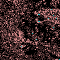}\\

\includegraphics[height=.2\textwidth, width = .18\textwidth ]{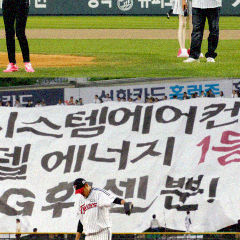}&
\includegraphics[height=.2\textwidth, width = .18\textwidth ]{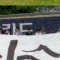}&
\includegraphics[height=.2\textwidth, width = .18\textwidth ]{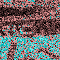}&
\includegraphics[height=.2\textwidth, width = .18\textwidth ]{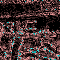}&
\includegraphics[height=.2\textwidth, width = .18\textwidth ]{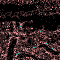}\\

\includegraphics[height=.2\textwidth, width = .18\textwidth ]{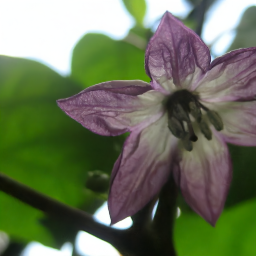}&
\includegraphics[height=.2\textwidth, width = .18\textwidth ]{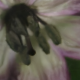}&
\includegraphics[height=.2\textwidth, width = .18\textwidth ]{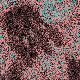}&
\includegraphics[height=.2\textwidth, width = .18\textwidth ]{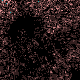}&
\includegraphics[height=.2\textwidth, width = .18\textwidth ]{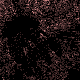}\\

Noisy input&Region of interest &Input error&BIPNET error&Our error\\

\end{tabular}

    \caption{Qualitative performance of our algorithm with respect to BIPNET~\cite{bipnet22} on the evaluation color dataset. Notice the excessive smoothing by BIPNET which removes the sharper features in comparison to the ground truth. Our network retains the details and removes the noise as well. Darker region corresponds to lesser error which indicates better denoising. It is to be noted that the error maps have been scaled 5 times for better visual understanding. }%
    \label{fig:Qual_an}
\end{figure*}

Exemplar qualitative results on individual images are shown in Figure~\ref{fig:Qual_an}

\begin{table}%[htbp]
  
\makebox[\textwidth][c]{
    \begin{tabular}{c|cc|cc}
\toprule
       & \multicolumn{2}{c|}{Color} & \multicolumn{2}{c}{Grayscale} \\
\midrule
model & lvl. 1  & lvl. 2  & lvl. 1  & lvl. 2 \\ \hline
KPN~\cite{mildenhall2018kpn} & 38.38 & 35.96 & 36.47 & 33.93 \\
BPN~\cite{BPN} & 40.16 & 37.08 & 38.18 & 35.42\\
MFIR-1~\cite{bhat2021deep} & 40.16 & 37.08 & 39.37 & 36.51\\
MFIR-2~\cite{bhat2021deep} & 42.21 & 39.13 & 39.37 & 36.51\\
BIPNET~\cite{bipnet22} & \textcolor{green!50!black}{42.28} & \textcolor{blue}{\textbf{40.20}} & \textcolor{green!50!black}{\textbf{41.26}} & \textcolor{blue}{\textbf{38.74}} \\
Ours & \textcolor{blue}{\textbf{42.49}} & \textcolor{green!50!black}{39.18} &  \textcolor{blue}{\textbf{41.35}} & \textcolor{green!50!black}{36.61} \\
\bottomrule
\end{tabular}%
}
\label{tab:quantitative}%
\vspace*{-0.4cm}
\caption{PSNR values of the evaluation grayscale burst dataset. Blue shows best results while green shows the second best results. \textit{lvl.} indicates the level of Gaussian noise added according to evaluation convention. }
\end{table}%

%%%%%%%%%%%%%%%%%%%

%------------------------------------------------------------------------- 
\subsection{Ablation Study}

\setlength{\tabcolsep}{6pt}
\begin{table}[htb]
\begin{center}

\label{table:ablation}
\begin{tabular}{lc}
\hline\noalign{\smallskip}
Model $\qquad\qquad$& \emph{lvl.1 gray} \\
\noalign{\smallskip}
\hline
\noalign{\smallskip}

without SGN pre-denoising & 40.81  \\
without RAFT alignment & 38.54 \\
without content-adaptive filtering & 41.27  \\

Ours -- complete pipeline & \textcolor{blue}{\textbf{41.35}} \\

\hline
\end{tabular}
\end{center}
\vspace*{-0.6cm}
\caption{Ablation study. Removing the individual parts of the pipeline and training the model from scratch reveals the importance of each component. Only by combining SGN-based pre-denosing, flow-based alignment and content-adpative filtering good performance in hight noise levels can be achieved.}
\label{table:ablation}
\end{table}

Since our module consists of several pretrained blocks and trainable sub-modules, we analyse the effectiveness of each of the components with the corresponding ablation in Table~\ref{table:ablation}. 
Here, we removed individual parts of the pipeline and trained the network from scratch. 
Removing all SGN blocks effectively suppresses pre-denoising of the input. 
Although the low-noise evaluation performs comparatively well, image quality deteriorates as the noise increases due to the lack of cleaner proposals at the initial stages. 
Without the alignment module, the final fusion step is impared and we see lower performance on all noise levels. Particularly \textit{lvl.1} is impacted. 
Finally, the cooperative content-adaptive filtering adds almost equally to the reconstruction quality of all noise levels levels. 

% Next, we remove the alignment module, which in turn further causes the deterioration of the network performance. This is due to the fact that due to lack of alignment of the random motions, the kernel filters are unable to determine effectively since a lot of pixel proposals go out of the limited filter kernel range. Next, we remove the kernel filters and we see a slight decrease in performance due to additional filtering in the later stages. Removal of SGN and Kernel filters altogether results in a jump in very low noise scenarios as the gausiian smoothing effect is gone and the network preserves the very high frequency signals, effectively showing a gain in performance only for low noise case. At the end, we show that in overall performance, our network with all the proposed module performs state of the art in terms of burst denoising, effectively showing the requirements of all the present blocks.  

%------------------------------------------------------------------------- RESULTS

\subsection{Qualitative Results}

In Figure~\ref{fig:Qual_an} we demonstrate the improved performance of our pipeline on a number of example images. Even for drastically different amount of noise our approach outperforms BIPNET~\cite{bipnet22} on every image. The denoised image is significantly closer to the ground truth result, as is evident from the error maps. %Additional results are shown as a supplement. 

\begin{figure*}
   \centering
\begin{tabular}{ccccc}

\includegraphics[height=.16\textwidth]{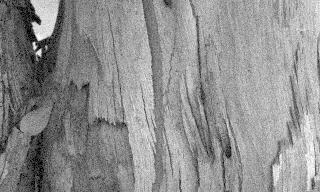}&
\includegraphics[height=.16\textwidth, width = .16\textwidth ]{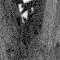}&
\includegraphics[height=.16\textwidth, width = .16\textwidth ]{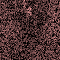}&
\includegraphics[height=.16\textwidth, width = .16\textwidth ]{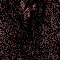}&
\includegraphics[height=.16\textwidth, width = .16\textwidth ]{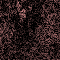}\\

Noisy input&Region of interest &Input error&BIPNET error&Our error\\

\end{tabular}

    \caption{Worst performing images in the test set. The excessive noise in the input combined with strong camera motion deteriorates the denoising performance. It is to be noted that the error maps have been scaled 2 times for better visual understanding. }
    \label{fig:failure} % I can do without the label too
\end{figure*}

A failure case is shown in Figure~\ref{fig:failure}. In this example, apart from the high motion, the image consists of sharp features which is preserved by our network which is not detected as noise.  %In the first example, the camera motion between the burst images was too high to obtain a proper alignment. similarly, in the second image the object motion resulted in too strong motion blur. 

\section{Conclusions}

We propose a deep-burst denoising model based on optical flow guided alignment and cooperative filtering. 
A well-established single image denoising module generates pre-denoised burst input images for two different assumed noise levels.
Alignment to the reference frame is performed using a state of the art optical flow network. 
Providing the original input burst and the pre-denoised stacks ensures the good performance of the optical flow alignment. 
Based on the aligned features and images a set of content-adaptive spatially-varying filter kernel is predicted to smooth each input image individually. 
A fusion block finally combines all intermediate results to the final denoised output. In the future, one can also compare the effect of state of the art optical flow based correspondence alignment on the quality of the burst image denosing.

Our approach yields state-of-the-art results across low noise levels on the standard benchmark data sets. Higher noise scenarios working on different pre-denoised images shows a comparable benefit. 

\section*{Acknowledgement}
This work has been partially funded by the Deutsche Forschungsgemeinschaft (DFG, German Research Foundation) under Germany’s Excellence Strategy - EXC number 2064/1 - project number 390727645 and SFB 1233 - project number 276693517. It was supported by the German Federal Ministry of Education and Research (BMBF): Tübingen AI Center, FKZ: 01IS18039A and Cyber Valley. The authors thank the International Max Planck Research School for Intelligent Systems (IMPRS-IS) for supporting Arijit Mallick.
% trigger a  just before the given reference

% Enable to reduce spacing between bibitems (source: https://tex.stackexchange.com/a/25774)
% \def\IEEEbibitemsep{0pt plus .5pt}
%\newpage
\bibliographystyle{IEEEtranN} % IEEEtranN is the natbib compatible bst file
\bibliography{egbib}

% Generated by IEEEtranN.bst, version: 1.14 (2015/08/26)
\begin{thebibliography}{39}
\providecommand{\natexlab}[1]{#1}
\providecommand{\url}[1]{#1}
\csname url@samestyle\endcsname
\providecommand{\newblock}{\relax}
\providecommand{\bibinfo}[2]{#2}
\providecommand{\BIBentrySTDinterwordspacing}{\spaceskip=0pt\relax}
\providecommand{\BIBentryALTinterwordstretchfactor}{4}
\providecommand{\BIBentryALTinterwordspacing}{\spaceskip=\fontdimen2\font plus
\BIBentryALTinterwordstretchfactor\fontdimen3\font minus
  \fontdimen4\font\relax}
\providecommand{\BIBforeignlanguage}[2]{{%
\expandafter\ifx\csname l@#1\endcsname\relax
\typeout{** WARNING: IEEEtranN.bst: No hyphenation pattern has been}%
\typeout{** loaded for the language `#1'. Using the pattern for}%
\typeout{** the default language instead.}%
\else
\language=\csname l@#1\endcsname
\fi
#2}}
\providecommand{\BIBdecl}{\relax}
\BIBdecl

\bibitem[Molini et~al.(2020)Molini, Valsesia, Fracastoro, and Magli]{sr2}
\BIBentryALTinterwordspacing
A.~B. Molini, D.~Valsesia, G.~Fracastoro, and E.~Magli, ``{DeepSUM}: Deep
  neural network for super-resolution of unregistered multitemporal images,''
  \emph{{IEEE} Transactions on Geoscience and Remote Sensing}, vol.~58, no.~5,
  pp. 3644--3656, may 2020. [Online]. Available:
  \url{https://doi.org/10.1109%2Ftgrs.2019.2959248}
\BIBentrySTDinterwordspacing

\bibitem[Kawulok et~al.(2020)Kawulok, Benecki, Piechaczek, Hrynczenko,
  Kostrzewa, and Nalepa]{sr3}
\BIBentryALTinterwordspacing
M.~Kawulok, P.~Benecki, S.~Piechaczek, K.~Hrynczenko, D.~Kostrzewa, and
  J.~Nalepa, ``Deep learning for multiple-image super-resolution,''
  \emph{{IEEE} Geoscience and Remote Sensing Letters}, vol.~17, no.~6, pp.
  1062--1066, jun 2020. [Online]. Available:
  \url{https://doi.org/10.1109%2Flgrs.2019.2940483}
\BIBentrySTDinterwordspacing

\bibitem[Liba et~al.(2019)Liba, Murthy, Tsai, Brooks, Xue, Karnad, He, Barron,
  Sharlet, Geiss, Hasinoff, Pritch, and Levoy]{Liba_2019}
\BIBentryALTinterwordspacing
O.~Liba, K.~Murthy, Y.-T. Tsai, T.~Brooks, T.~Xue, N.~Karnad, Q.~He, J.~T.
  Barron, D.~Sharlet, R.~Geiss, S.~W. Hasinoff, Y.~Pritch, and M.~Levoy,
  ``Handheld mobile photography in very low light,'' \emph{{ACM} Transactions
  on Graphics}, vol.~38, no.~6, pp. 1--16, dec 2019. [Online]. Available:
  \url{https://doi.org/10.1145%2F3355089.3356508}
\BIBentrySTDinterwordspacing

\bibitem[Hasinoff et~al.(2016)Hasinoff, Sharlet, Geiss, Adams, Barron, Kainz,
  Chen, and Levoy]{burst_p1}
\BIBentryALTinterwordspacing
S.~W. Hasinoff, D.~Sharlet, R.~Geiss, A.~Adams, J.~T. Barron, F.~Kainz,
  J.~Chen, and M.~Levoy, ``Burst photography for high dynamic range and
  low-light imaging on mobile cameras,'' \emph{ACM Trans. Graph.}, vol.~35,
  no.~6, nov 2016. [Online]. Available:
  \url{https://doi.org/10.1145/2980179.2980254}
\BIBentrySTDinterwordspacing

\bibitem[Bhat et~al.(2021{\natexlab{a}})Bhat, Danelljan, Van~Gool, and
  Timofte]{bhat1}
\BIBentryALTinterwordspacing
G.~Bhat, M.~Danelljan, L.~Van~Gool, and R.~Timofte, ``Deep burst
  super-resolution,'' 2021. [Online]. Available:
  \url{https://arxiv.org/abs/2101.10997}
\BIBentrySTDinterwordspacing

\bibitem[Wronski et~al.(2019)Wronski, Garcia-Dorado, Ernst, Kelly, Krainin,
  Liang, Levoy, and Milanfar]{burst_sr1}
\BIBentryALTinterwordspacing
B.~Wronski, I.~Garcia-Dorado, M.~Ernst, D.~Kelly, M.~Krainin, C.-K. Liang,
  M.~Levoy, and P.~Milanfar, ``Handheld multi-frame super-resolution,''
  \emph{ACM Trans. Graph.}, vol.~38, no.~4, jul 2019. [Online]. Available:
  \url{https://doi.org/10.1145/3306346.3323024}
\BIBentrySTDinterwordspacing

\bibitem[Valsesia and Magli(2022)]{valsesia2021piunet}
D.~Valsesia and E.~Magli, ``Permutation invariance and uncertainty in
  multitemporal image super-resolution,'' \emph{IEEE Transactions on Geoscience
  and Remote Sensing}, vol.~60, pp. 1--12, 2022.

\bibitem[Deudon et~al.(2020)Deudon, Kalaitzis, Goytom, Arefin, Lin, Sankaran,
  Michalski, Kahou, Cornebise, and Bengio]{highresnet2020}
\BIBentryALTinterwordspacing
M.~Deudon, A.~Kalaitzis, I.~Goytom, M.~R. Arefin, Z.~Lin, K.~Sankaran,
  V.~Michalski, S.~E. Kahou, J.~Cornebise, and Y.~Bengio, ``Highres-net:
  Recursive fusion for multi-frame super-resolution of satellite imagery,''
  2020. [Online]. Available: \url{https://arxiv.org/abs/2002.06460}
\BIBentrySTDinterwordspacing

\bibitem[Godard et~al.(2018)Godard, Matzen, and Uyttendaele]{Godard_2018_ECCV}
C.~Godard, K.~Matzen, and M.~Uyttendaele, ``Deep burst denoising,'' in
  \emph{Proceedings of the European Conference on Computer Vision (ECCV)},
  September 2018.

\bibitem[Bhat et~al.(2021{\natexlab{b}})Bhat, Danelljan, Yu, Gool, and
  Timofte]{bhat2021deep}
G.~Bhat, M.~Danelljan, F.~Yu, L.~V. Gool, and R.~Timofte, ``Deep
  reparametrization of multi-frame super-resolution and denoising,'' 2021.

\bibitem[Gu et~al.(2019)Gu, Li, Gool, and Timofte]{gu2019self}
S.~Gu, Y.~Li, L.~V. Gool, and R.~Timofte, ``Self-guided network for fast image
  denoising,'' in \emph{Proceedings of the IEEE/CVF International Conference on
  Computer Vision}, 2019, pp. 2511--2520.

\bibitem[Teed and Deng(2020)]{teed2020raft}
Z.~Teed and J.~Deng, ``Raft: Recurrent all-pairs field transforms for optical
  flow,'' in \emph{European conference on computer vision}.\hskip 1em plus
  0.5em minus 0.4em\relax Springer, 2020, pp. 402--419.

\bibitem[Dabov et~al.(2007{\natexlab{a}})Dabov, Foi, Katkovnik, and
  Egiazarian]{BM3D}
K.~Dabov, A.~Foi, V.~Katkovnik, and K.~Egiazarian, ``Image denoising by sparse
  3d transform-domain collaborative filtering,'' 2007.

\bibitem[Xie et~al.(2012)Xie, Xu, and Chen]{inpainting}
J.~Xie, L.~Xu, and E.~Chen, ``Image denoising and inpainting with deep neural
  networks,'' in \emph{Proceedings of the 25th International Conference on
  Neural Information Processing Systems - Volume 1}, ser. NIPS'12.\hskip 1em
  plus 0.5em minus 0.4em\relax Red Hook, NY, USA: Curran Associates Inc., 2012,
  p. 341–349.

\bibitem[Aharon et~al.(2006)Aharon, Elad, and Bruckstein]{KSVD}
M.~Aharon, M.~Elad, and A.~Bruckstein, ``K-svd: An algorithm for designing
  overcomplete dictionaries for sparse representation,'' \emph{IEEE
  Transactions on Signal Processing}, vol.~54, no.~11, pp. 4311--4322, 2006.

\bibitem[Schmidt and Roth(2014)]{fields}
U.~Schmidt and S.~Roth, ``Shrinkage fields for effective image restoration,''
  in \emph{2014 IEEE Conference on Computer Vision and Pattern Recognition},
  2014, pp. 2774--2781.

\bibitem[Zhang et~al.(2017)Zhang, Zuo, Chen, Meng, and Zhang]{resD}
\BIBentryALTinterwordspacing
K.~Zhang, W.~Zuo, Y.~Chen, D.~Meng, and L.~Zhang, ``Beyond a gaussian denoiser:
  Residual learning of deep {CNN} for image denoising,'' \emph{{IEEE}
  Transactions on Image Processing}, vol.~26, no.~7, pp. 3142--3155, jul 2017.
  [Online]. Available: \url{https://doi.org/10.1109%2Ftip.2017.2662206}
\BIBentrySTDinterwordspacing

\bibitem[Mao et~al.(2016)Mao, Shen, and Yang]{imRestore}
\BIBentryALTinterwordspacing
X.-J. Mao, C.~Shen, and Y.-B. Yang, ``Image restoration using very deep
  convolutional encoder-decoder networks with symmetric skip connections,''
  2016. [Online]. Available: \url{https://arxiv.org/abs/1603.09056}
\BIBentrySTDinterwordspacing

\bibitem[Tai et~al.(2017)Tai, Yang, Liu, and Xu]{Memnet}
\BIBentryALTinterwordspacing
Y.~Tai, J.~Yang, X.~Liu, and C.~Xu, ``Memnet: A persistent memory network for
  image restoration,'' 2017. [Online]. Available:
  \url{https://arxiv.org/abs/1708.02209}
\BIBentrySTDinterwordspacing

\bibitem[Chen et~al.(2020)Chen, Wang, Guo, Xu, Deng, Liu, Ma, Xu, Xu, and
  Gao]{Itrans}
\BIBentryALTinterwordspacing
H.~Chen, Y.~Wang, T.~Guo, C.~Xu, Y.~Deng, Z.~Liu, S.~Ma, C.~Xu, C.~Xu, and
  W.~Gao, ``Pre-trained image processing transformer,'' 2020. [Online].
  Available: \url{https://arxiv.org/abs/2012.00364}
\BIBentrySTDinterwordspacing

\bibitem[Wieschollek et~al.(2017)Wieschollek, Hirsch, Scholkopf, and
  Lensch]{Wieschollek_2017_ICCV}
P.~Wieschollek, M.~Hirsch, B.~Scholkopf, and H.~P.~A. Lensch, ``Learning blind
  motion deblurring,'' in \emph{Proceedings of the IEEE International
  Conference on Computer Vision (ICCV)}, Oct 2017.

\bibitem[Mildenhall et~al.(2018)Mildenhall, Barron, Chen, Sharlet, Ng, and
  Carroll]{mildenhall2018kpn}
B.~Mildenhall, J.~T. Barron, J.~Chen, D.~Sharlet, R.~Ng, and R.~Carroll,
  ``Burst denoising with kernel prediction networks,'' in \emph{IEEE Conference
  on Computer Vision and Pattern Recognition (CVPR)}, 2018.

\bibitem[Cho et~al.(2021)Cho, Son, and Kim]{mkpn}
W.~Cho, S.~Son, and D.-S. Kim, ``Weighted multi-kernel prediction network for
  burst image super-resolution,'' in \emph{2021 IEEE/CVF Conference on Computer
  Vision and Pattern Recognition Workshops (CVPRW)}, 2021, pp. 404--413.

\bibitem[Dudhane et~al.(2022)Dudhane, Zamir, Khan, Khan, and Yang]{bipnet22}
A.~Dudhane, S.~W. Zamir, S.~Khan, F.~S. Khan, and M.-H. Yang, ``Burst image
  restoration and enhancement,'' in \emph{2022 IEEE/CVF Conference on Computer
  Vision and Pattern Recognition (CVPR)}, 2022, pp. 5749--5758.

\bibitem[Tico(2008)]{mfid}
M.~Tico, ``Multi-frame image denoising and stabilization,'' in \emph{2008 16th
  European Signal Processing Conference}, 2008, pp. 1--4.

\bibitem[Maggioni et~al.(2012)Maggioni, Boracchi, Foi, and Egiazarian]{VBM4D}
M.~Maggioni, G.~Boracchi, A.~Foi, and K.~Egiazarian, ``Video denoising,
  deblocking, and enhancement through separable 4-d nonlocal spatiotemporal
  transforms,'' \emph{IEEE Transactions on Image Processing}, vol.~21, no.~9,
  pp. 3952--3966, 2012.

\bibitem[Dabov et~al.(2007{\natexlab{b}})Dabov, Foi, and Egiazarian]{VBM3D}
K.~Dabov, A.~Foi, and K.~Egiazarian, ``Video denoising by sparse 3d
  transform-domain collaborative filtering,'' in \emph{2007 15th European
  Signal Processing Conference}, 2007, pp. 145--149.

\bibitem[Bruhn et~al.(2005)Bruhn, Weickert, and Schn{\"o}rr]{bruhn2005lucas}
A.~Bruhn, J.~Weickert, and C.~Schn{\"o}rr, ``Lucas/kanade meets horn/schunck:
  Combining local and global optic flow methods,'' \emph{International journal
  of computer vision}, vol.~61, no.~3, pp. 211--231, 2005.

\bibitem[Buades et~al.(2016)Buades, Lisani, and Miladinović]{buades2016}
A.~Buades, J.-L. Lisani, and M.~Miladinović, ``Patch-based video denoising
  with optical flow estimation,'' \emph{IEEE Transactions on Image Processing},
  vol.~25, no.~6, pp. 2573--2586, 2016.

\bibitem[Dosovitskiy et~al.(2015)Dosovitskiy, Fischer, Ilg, Hausser, Hazirbas,
  Golkov, Van Der~Smagt, Cremers, and Brox]{dosovitskiy2015flownet}
A.~Dosovitskiy, P.~Fischer, E.~Ilg, P.~Hausser, C.~Hazirbas, V.~Golkov, P.~Van
  Der~Smagt, D.~Cremers, and T.~Brox, ``Flownet: Learning optical flow with
  convolutional networks,'' in \emph{Proceedings of the IEEE international
  conference on computer vision}, 2015, pp. 2758--2766.

\bibitem[Ilg et~al.(2017)Ilg, Mayer, Saikia, Keuper, Dosovitskiy, and
  Brox]{ilg2017flownet}
E.~Ilg, N.~Mayer, T.~Saikia, M.~Keuper, A.~Dosovitskiy, and T.~Brox, ``Flownet
  2.0: Evolution of optical flow estimation with deep networks,'' in
  \emph{Proceedings of the IEEE conference on computer vision and pattern
  recognition}, 2017, pp. 2462--2470.

\bibitem[Huang et~al.(2022)Huang, Shi, Zhang, Wang, Cheung, Qin, Dai, and
  Li]{huang2022flowformer}
Z.~Huang, X.~Shi, C.~Zhang, Q.~Wang, K.~C. Cheung, H.~Qin, J.~Dai, and H.~Li,
  ``Flowformer: A transformer architecture for optical flow,'' \emph{arXiv
  preprint arXiv:2203.16194}, 2022.

\bibitem[Butler et~al.(2012)Butler, Wulff, Stanley, and
  Black]{Butler:ECCV:2012}
D.~J. Butler, J.~Wulff, G.~B. Stanley, and M.~J. Black, ``A naturalistic open
  source movie for optical flow evaluation,'' in \emph{European Conf. on
  Computer Vision (ECCV)}, ser. Part IV, LNCS 7577, {A. Fitzgibbon et al.
  (Eds.)}, Ed.\hskip 1em plus 0.5em minus 0.4em\relax Springer-Verlag, Oct.
  2012, pp. 611--625.

\bibitem[Paszke et~al.(2019)Paszke, Gross, Massa, Lerer, Bradbury, Chanan,
  Killeen, Lin, Gimelshein, Antiga, Desmaison, Kopf, Yang, DeVito, Raison,
  Tejani, Chilamkurthy, Steiner, Fang, Bai, and Chintala]{NEURIPS2019_9015}
\BIBentryALTinterwordspacing
A.~Paszke, S.~Gross, F.~Massa, A.~Lerer, J.~Bradbury, G.~Chanan, T.~Killeen,
  Z.~Lin, N.~Gimelshein, L.~Antiga, A.~Desmaison, A.~Kopf, E.~Yang, Z.~DeVito,
  M.~Raison, A.~Tejani, S.~Chilamkurthy, B.~Steiner, L.~Fang, J.~Bai, and
  S.~Chintala, ``Pytorch: An imperative style, high-performance deep learning
  library,'' in \emph{Advances in Neural Information Processing Systems 32},
  H.~Wallach, H.~Larochelle, A.~Beygelzimer, F.~d\textquotesingle
  Alch\'{e}-Buc, E.~Fox, and R.~Garnett, Eds.\hskip 1em plus 0.5em minus
  0.4em\relax Curran Associates, Inc., 2019, pp. 8024--8035. [Online].
  Available:
  \url{http://papers.neurips.cc/paper/9015-pytorch-an-imperative-style-high-performance-deep-learning-library.pdf}
\BIBentrySTDinterwordspacing

\bibitem[Kingma and Ba(2014)]{kingma2014adam}
D.~P. Kingma and J.~Ba, ``Adam: A method for stochastic optimization,''
  \emph{arXiv preprint arXiv:1412.6980}, 2014.

\bibitem[ZHAO(2019)]{SGNgit}
\BIBentryALTinterwordspacing
Y.~ZHAO, ``{PyTorch implementation of Self Guided Network (ICCV)},'' 2019.
  [Online]. Available:
  \url{https://github.com/zhaoyuzhi/Self-Guided-Network-for-Fast-Image-Denoising}
\BIBentrySTDinterwordspacing

\bibitem[Krasin et~al.(2017)Krasin, Duerig, Alldrin, Ferrari, Abu-El-Haija,
  Kuznetsova, Rom, Uijlings, Popov, Kamali, Malloci, Pont-Tuset, Veit,
  Belongie, Gomes, Gupta, Sun, Chechik, Cai, Feng, Narayanan, and
  Murphy]{OpenImages2}
I.~Krasin, T.~Duerig, N.~Alldrin, V.~Ferrari, S.~Abu-El-Haija, A.~Kuznetsova,
  H.~Rom, J.~Uijlings, S.~Popov, S.~Kamali, M.~Malloci, J.~Pont-Tuset, A.~Veit,
  S.~Belongie, V.~Gomes, A.~Gupta, C.~Sun, G.~Chechik, D.~Cai, Z.~Feng,
  D.~Narayanan, and K.~Murphy, ``Openimages: A public dataset for large-scale
  multi-label and multi-class image classification.'' \emph{Dataset available
  from https://storage.googleapis.com/openimages/web/index.html}, 2017.

\bibitem[Healey and Kondepudy(1994)]{hetero}
G.~Healey and R.~Kondepudy, ``Radiometric ccd camera calibration and noise
  estimation,'' \emph{IEEE Transactions on Pattern Analysis and Machine
  Intelligence}, vol.~16, no.~3, pp. 267--276, 1994.

\bibitem[Xia et~al.(2020)Xia, Perazzi, Gharbi, Sunkavalli, and
  Chakrabarti]{BPN}
Z.~Xia, F.~Perazzi, M.~Gharbi, K.~Sunkavalli, and A.~Chakrabarti, ``Basis
  prediction networks for effective burst denoising with large kernels,'' in
  \emph{2020 IEEE/CVF Conference on Computer Vision and Pattern Recognition
  (CVPR)}, 2020, pp. 11\,841--11\,850.

\end{thebibliography}

\end{document}